\documentclass{article}

\usepackage{aimc2022}

\usepackage[utf8]{inputenc} 
\usepackage[T1]{fontenc}    
\usepackage{hyperref}       
\usepackage{url}            
\usepackage{booktabs}       
\usepackage{amsfonts}       
\usepackage{nicefrac}       
\usepackage{microtype}      
\usepackage{graphicx}       
\usepackage{bm}

\newcommand{\supp}[2][]{\mathrm{supp}_{#1} \ #2}
\newcommand{\bx}{\mathbf{x}}
\newcommand{\bX}{\mathbf{X}}
\newcommand{\by}{\mathbf{y}}
\newcommand{\bz}{\mathbf{z}}
\newcommand{\btheta}{\bm{\theta}}
\newcommand{\bepsilon}{\bm{\epsilon}}
\newcommand{\bphi}{\bm{\phi}}

\newcommand{\bxi}{\bm{\xi}}
\newcommand{\bomega}{\bm{\omega}}

\newcommand{\comment}[1]{}

\title{Challenges in creative generative models for music: a divergence maximization perspective}

\author{%
  Axel Chemla--Romeu-Santos\\
  IRCAM\\
  4, place Igor Stravinsky\\
  75004, Paris, France\\
  \texttt{chemla@ircam.fr} \\
  \And
  Philippe Esling\\
  IRCAM\\
  4, place Igor Stravinsky\\
  75004, Paris, France\\
  \texttt{esling@ircam.fr} \\
}

\begin{document}

\maketitle

\begin{abstract}
The development of generative Machine Learning (ML) models in creative practices, enabled by the recent improvements in usability and availability of pre-trained models, is raising more and more interest among artists, practitioners and performers. Yet, the introduction of such techniques in artistic domains also revealed multiple limitations that escape current evaluation methods used by scientists. Notably, most models are still unable to generate content that lay outside of the domain defined by the training dataset. In this paper, we propose an alternative prospective framework, starting from a new general formulation of ML objectives, that we derive to delineate possible implications and solutions that already exist in the ML literature (notably for the audio and musical domain). We also discuss existing relations between generative models and computational creativity and how our framework could help address the lack of creativity in existing models. 
\end{abstract}
\section{Introduction}
During the last decade, important efforts in Machine Learning (ML) have been dedicated to  \textit{generative models} \cite{bishop2006pattern}, mainly caused by the extremely high number of dimensions of the input data. These efforts have yielded a profuse diversity of approaches, that can be in certain domains (notably image) almost on par with established creation tools for both expert and non-expert users alike. Yet, especially in audio and music generation, the gap between the computational power of recent machine learning systems and their actual use in creative works is still surprisingly large. Indeed, most of the researches done so far have focused on the accuracy and quality aspects of these models. Hence, this shadowed important aspects for their creative use: the difficulty of evaluating the generation of novel materials, the use of losses that can hinder the diversity of the generated content, and the lack of investment in analyzing how these models behave across various computational co-creativity tasks. We think that these shortcomings bias these models to generate "more of the same" content \citet{mccormack2020design}, causing two complementary issues. First, fostering reconstruction instead of extrapolation prevent ML methods to thrive a real interest of artists that would like to push these systems to their limits. Secondly, leaving the behavior of such systems in extensional cases unexplored can limit their integration in co-creative setups (and human-computer networks in general).
While recent approaches tried to extend and investigate the generative abilities of these models (such as \textit{active divergence} \citet{broad2021active}), we believe that the global framework used for training generative models must also be questioned. Furthermore, these approaches (that we call \textit{extensional}) have hardly been tried in music generation (audio or symbolic), and then still totally absent from musical composition/performance. Rather than discussing creativity in general, we will focus on specific points of ML-based generation models. Here, we try to delineate general objectives for extensional cases to stimulate the creation of content that diverge from existing data ; in other words, encouraging the model to \textit{extrapolate} the original data distribution. We also discuss how different computational creativity notions could be integrated in this new objective, and if they can be used for evaluation or optimization purposes. Finally, we underline current limitations and perspectives for the creative application of generative models, and why such developments are mandatory for not only artistic but also societal purposes.


\section{Context}

\subsection{Generative models}
Generative models learn a dataset of $N$ examples $\bX = \{ \bx_1 ... \bx_N \}$, supposedly following an unknown probability distribution $p(\bx)$ ($\bx \in \mathcal{X}$) by using a model $p_{\btheta}(\bx)$, whose parameters $\btheta$ are updated using convex optimization. However, as such models would only allow random sampling, generative models are usually \textit{conditioned} on another set of variables, that we denote as $\mathcal{C}$, that can be used to control or contextualize the generation. This objective can be written 
\begin{equation} 
\label{eq:div_min}
\min_{\btheta} \mathcal{D}[p_{\btheta}(\bx | \mathcal{C})\Vert p(\bx)],
\end{equation}
where $\mathcal{D}[\cdot \Vert \cdot]$ is a divergence measuring the difference between the two distributions. Within this framework, $\bx$ can equivalently be audio waveforms, spectrograms, piano-rolls, or any other audio/musical representation. The diverse models in the literature will then differ in the chosen divergence $\mathcal{D}$, the conditioning variables $\mathcal{C}$ (that can be given or inferred), and the training method. For instance, while some models directly model the full distribution $p(\bx)$, auto-regressive approaches rather model the conditional distribution $p(\bx_{i+1}|\bx_{i})$, which can be useful for high-dimensional data or sequence modelling. Regarding the conditioning variables $\mathcal{C}$, these can be direct information regarding the data, usually denoted as $\by$, that can be used to control the generation during inference. Alternatively, $\mathcal{C}$ can also include additional stochastic variables $\bz$, called \textit{latent variables}. These variables can either be inferred from the data with an \textit{inference} model, as in \textit{Variational Auto-Encoders} (VAEs) \citet{kingma2013auto}, or sampled during training, as in \textit{Generative Adversarial Networks} (GANs) \citet{goodfellow2014generative}. Finally, the choice of the divergence $\mathcal{D}$ is also decisive, and can either be a fixed training criteria (e.g. log-likelihood, mean-squared error) or can also be optimized during training, as in adversarial setups \citet{arjovsky2017wasserstein}.
The application of these models to audio and musical data has been extensively studied in the last decade. Encoding-decoding architectures have been leveraged for both symbol-to-symbol \citet{roberts2018hierarchical}, symbol-to-audio \citet{defossez2018sing, dhariwal2020jukebox}, audio-to-symbol \citet{hawthorne2017onsets} and audio generation \citet{esling2018generative, engel2017neural, engel2020ddsp, caillon2021rave}. Similarly, adversarial architectures have been used for both symbolic \citet{muhamed2020transformer, zhang2021implement, greshler2021catch} or raw audio generation \citet{engel2019gansynth, broek2021mp3net}. Due to the complexity of this task, alternative models such as normalizing flows \citet{prenger2019waveglow} (NFs) or diffusion models \citet{kong2020diffwave} (DFs) have also been successfully used, both also involving latent representations $\bz$. For more comprehensive surveys of these research domains, see \citet{briot2017deep, shi2021survey}.

\subsection{Evaluation methods} 
Despite the profusion of research in ML models for sound and music generation, the development of evaluation methods for these approaches has been much more scarcely addressed. Indeed, evaluating generative abilities of a model without a comparative reference is a very complex task, such that most of them are validated through reconstruction scores. Besides, another overarching goal in machine learning evaluation lies in \textit{generalization}, which aims to assess how much the system is able to extend the learned features to unseen data. However, generalization itself is also an elusive concept, as we can define it either as their resilience to outliers, or their ability to extrapolate towards an out-of-domain distribution. To this end, some validation methods ground their evaluation on the ability of the model to represent all the underlying classes of a dataset, resorting to external label information \citet{theis2015note}. Alternatively, other methods aim to provide domain-dependant evaluations using \textit{perceptive} attributes \citet{manocha2020differentiable}. However, these task-specific evaluation methods only measure the capacity of the model to generate existing content, which discards evaluating their potential in computationally creative or human-computer co-creative setups. Indeed, such evaluations are difficult to quantify as there is no straightforward notion of optimality and are also inherently context-dependant \citet{pasquier2017introduction}. In this paper, we rely on the observation that current models are developed by focusing on the idea of \textit{typicality} rather than \textit{novelty} \cite{colton2012computational}, preventing to promote these models for creative purposes \citet{diedrich2015creative}. Hence, we propose to take a different perspective and study how novelty objectives could be integrated directly during training, hence allowing to shift from \textit{training database} to \textit{inspiring sets}. 

\section{Formulation attempts and insights}
\label{sec:formulation}
As aforementioned, obj.~\ref{eq:div_min} implicitly prevents ML-based generative models from going away from the reference data distribution, and then to be pushed towards extensional cases. In this section, we will first propose a new objective that would enforce the model to innovate directly in the data domain, as well as analysing some of its intrinsic limitations. Then, we will try to extend it to other variables of the models. Finally, we will propose a meta-learning version of our objective, that can leverage a set of pre-trained generative models to control specific aspects of the targeted divergence. 

\begin{figure}[t]
  \label{fig:supports}
  \begin{center}
  \includegraphics[width=0.9\textwidth]{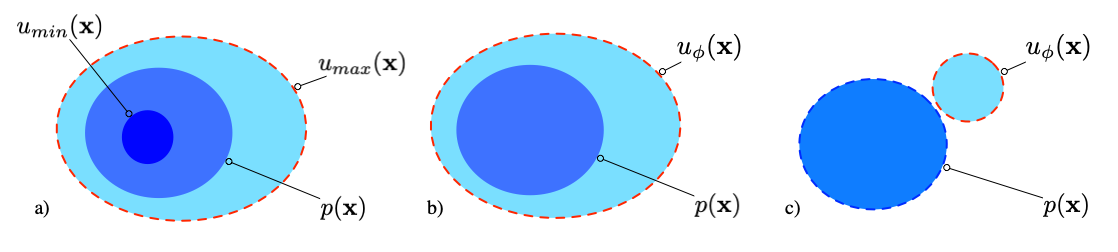}
      \caption{(a) The \textit{divergence ambiguity}, where $u_{min}(\bx)$ can only generate a subset of $p(\bx)$, while $u_{max}(\bx)$ correctly extrapolates $p(\bx)$. (b) a \textit{total extrapolation} case, where $\supp{u_{\bphi}(\bx)} \cap \supp{p(\bx)} = \supp{p(\bx)}$. (c) a \textit{total transfer} case, where $\supp{u_{\bphi}(\bx)} \cap \supp{p(\bx)} = \emptyset$.}
  \end{center}
\end{figure}

\subsection{Formalizing extrapolation as a divergence maximization}
\label{sec:div_max}
In order to open ML generative models to extensional cases, we propose to decline obj.~\ref{eq:div_min} to rather optimize a \textit{divergence maximization}, that we would formulate
\begin{equation} 
\label{eq:div_max}
\max_{\bphi} \mathcal{D}[u_{\bphi}(\bx | \mathcal{C}) \Vert p(\bx)]
\end{equation}
where we want our target model $u_{\bphi}(\bx | \mathcal{C})$, parameterized by $\bphi$, to diverge from the data distribution $p(\bx)$. While the optimization of this objective would clearly push the generative model away from the original distribution, it can also lead to \textit{catastrophic destruction}, where the maximised data would be totally uncorrelated to the target (e.g. white noise). Hence, we need additional criteria for guiding $u_{\bphi}(\bx | \mathcal{C})$ towards non-degenerated distributions. While bounding obj.~\ref{eq:div_max} could hinder the model from escaping too much the data distribution, we could obtain a finer description by rather comparing the support of $u_\phi(\bx | \mathcal{C})$ and an approximation of the support of the training distribution. Suppose we define a measure for the support of distribution $\supp{p(\bx)}$. This can be approximated by the convex hull of the training dataset $\supp{p(\bx)} = \mathrm{ConvexHull}(\bx)$ with $\bx \sim p(\bx)$. 
In our specific case, eq.~\ref{eq:div_max} can be equivalently optimized by two different distributions $u^{min}_{\bphi}(\bx)$ and $u^{max}_{\bphi}(\bx)$, such that $\supp{u^{min}_{\bphi}(\bx)} \subseteq \supp{p(\bx)} \subseteq \supp{u^{max}_{\bphi}(\bx)}$, as depicted in fig.~1a. However, the first distribution $\supp{u^{min}_{\bphi}(\bx)}$ leads to an inapt generative distribution (e.g. generating only a few examples of the dataset), while the second might correspond to a proper extrapolation of the generative distribution $p(\bx)$. Hence, the divergence eq.~\ref{eq:div_max} is not sufficient neither to provide a sufficient diversity, nor an adequate behavior on extensional cases. \\
Looking deeper into distribution supports allows to differentiate between different objectives in ML approaches. These could be integrated inside training, given an approximation of the support $\supp{p(\bx)}$. We can start by identifying two extreme cases: if $\supp{u_{\bphi}(\bx)} \cap \supp{p(\bx)} = \emptyset$, we are in a \textit{total transfer} setup (note that $\mathcal{D}$ could tend to $+\infty$). Conversely, if $\supp{u_{\bphi}(\bx)} \cap \supp{p(\bx)} = \supp{p(\bx)}$, we are in an \textit{total extrapolation} setup (see fig.1b-c). We can observe that, while these objectives are opposite, intermediate cases are still possible, and could be evaluated directly if provided evaluation measures such as precision / recall \citet{borji2022pros}. Then, we could control at which point the target distribution $u_{\bphi}(\bx)$ should differ from $p(\bx)$, and push the model to extrapolation regimes by simultaneously maximizing obj.~\ref{eq:div_max} and enforcing $\supp{u_{\bphi}(\bx)} \cap \supp{p(\bx)} = \supp{p(\bx)}$. While this would enforce the model to an extensional setup, we can note that additional criterion does not prevent catastrophic destruction.

\subsection{Generalizing the maximum divergence} 
\label{sec:gen_max_div}
The maximum divergence objective obj.~\ref{eq:div_max} can also be extended to other domains (besides the data $\mathcal{X}$). Notably, we can target both latent domain $\bz \in \mathcal{Z}$ and parameter domain $\bphi \in \Theta$. Here, we will consider a pre-trained latent parametric model $p_{\btheta}(\bx, \bz)$ with tractable inference distribution $p_{\btheta}(\bz | \bx)$ and aggregated posterior distribution $p_{\btheta}(\bz) = \int_\mathcal{X} p_{\btheta}(\bz | \bx) d\bx$. Defining our divergent distribution $u_{\bphi}(\bz)$ in the latent domain, we can use obj.~\ref{eq:div_max} to perform \textit{latent exploration} of the original latent space. We could then consider $u_{\bphi}(\bz)$ as a divergent prior, seeking positions that do not correspond to any item of the training dataset of $p_{\btheta}(\bx, \bz)$. 
Similarly, the maximum divergence objective can be extended to the parameter space. Considering a target generative model $u_{\bphi}(\bz | \bx)$, with the same parameter space $\Theta$ as $p_{\btheta}(\bx, \bz)$, we can also push the posterior distribution $u(\bphi | \bx, \bz)$ away from the original $p(\btheta | \bx, \bz)$. By targeting specific subsets of the parameter space $\Theta$ (encoding / decoding parameters, or specific layers), we can hence promote our target model to diverge in specific information layers (see sec.~\ref{sec:meta-max}). Therefore, we can extend our original objective eq.~\ref{eq:div_max} as follows :
\begin{itemize}
    \item $\max_{\bphi} \mathcal{D}[u_{\bphi}(\bx) \Vert p_{\btheta}(\bx)]$ : deviation in data domain (domain transformation)
    \item $\max_{\bphi} \mathcal{D}[u_{\bphi}(\mathbf{z}) \Vert p_{\btheta}(\mathbf{z})]$ :  deviation in latent domain (latent exploration)
    \item $\max_{\bphi} \mathcal{D}[u(\mathbf{\bphi}|\bx, \bz) \Vert p(\mathbf{\btheta}|\bx, \bz)]$ :  deviation in parameter domain
\end{itemize}
Considering joint models $u(\bx, \bz, \bphi)$ and $p(\bx, \bz, \btheta)$, whose conditioning corresponds to specific architectures, these divergences can be made independent or optimized together, providing different ways to control how our target model diverge from $p(\bx, \bz, \btheta)$. Moreover, this formulation encompasses many attempts done in the literature to diverge generative models from existing data : latent space exploration \citet{kegl2018spurious}, neural style transfer \citet{jing2019neural}, or loss hacking \citet{broad2020amplifying}.\\

\textbf{Global vs point-wise divergences.} In the case of latent generative models, where an output is produced by a generator function $g_{\bphi}(\bz)$, some models (as VAEs) compare point-wise evaluations $\mathcal{D}[g_{\bphi}(\bz_i) \Vert p(\bx_{i})]$, while other models such as GANs rather perform aggregated evaluations $\mathcal{D}[g_{\bphi}(\bz) \Vert p(\bx)]$. In the first case, point-wise evaluations act as a \textit{reconstruction loss}, favoring isolated latent positions $\bz_i$, while in the second case the divergence rather acts as a \textit{domain critic}, favoring information sharing between examples \citet{genevay2017gan, bousquet2017optimal}. These two types of divergences produce different kinds of artifacts, and must be then carefully designed when looking for a specific type of \textit{novelty} (see sec.~\ref{sec:cc}). Reversely, maximizing whether aggregated or point-wise divergences would have a significant impact on the model's outcome : in the first case, a single outcome $\bx_i$ would be pushed away from the entire distribution $p(\bx)$, while in the second case it could simply maximize obj.~\ref{eq:div_max} by matching another example $\bx_j$ of the dataset. Finally, more subtle grouping strategies could create intermediate steps between these two extreme cases, and hence be used to control the nature of the obtained artifacts (see sec.~\ref{sec:meta-max}).

\subsection{Formalizing extrapolation as a meta-learning problem}
\label{sec:meta-max}
Here, we consider that any given training dataset (with underlying density $p^{(i)}(\bx)$) is actually only a subset of the collection of all possible datasets of a given domain. Hence, the whole domain itself could be defined as a mixture model $p(\bx) = \sum_j \omega_j p^{(j)}(\bx)$, with $\sum_j \omega_j = 1$. Hence, we can extend obj.~\ref{eq:div_max} to a more scientifically sound \textit{maximum divergence objective}:
\begin{equation}
\label{eq:max_div_2}
\max_{\bphi} \frac{\mathcal{D}\left[  u_{\bphi}(\bx)  \parallel p^{(i)}(\bx) \right]}{\mathcal{D} \left[ u_{\bphi}(\bx)  \parallel p(\bx) \right ]} = \max_{\bphi} \frac{\mathcal{D}\left[ u_{\bphi}(\bx) \parallel p^{(i)}(\bx) \right ]}{\mathcal{D}\left[  u_{\bphi}(\bx) \parallel \sum_{j} \omega_{j}p^{(j)}(\bx)  \right ]}
\end{equation}
Therefore, this objective aims to maximize the divergence of distribution $u_{\bphi}(\bx)$ to a given example set $p^{(i)}(\bx)$ (such that $\mathcal{D}[u_{\bphi}(\bx) \Vert p^{(i)}(\bx)] \rightarrow \infty$), while simultaneously remaining coherent to the mixture of all possible datasets of a given domain $p(\bx)$ (with $\mathcal{D}[u_{\bphi}(\bx) \Vert p(\bx)] \rightarrow 0$). 

\textbf{Meta-learning approach.} Now, let us consider a collection of datasets $\mathcal{T} = \{ \bX^{(i)} \}_{i \in \mathbb{N}}$. This partitioning can be performed randomly (until the atomic case, where each dataset contains a unique element), or by gathering examples with shared attributes (e.g instrument or style). Then, we can make the distribution $u_{\bphi}(\bx)$ diverge from a single dataset of underlying probability $p^{(i)}(\bx)$ by optimizing 
\begin{equation}
\label{eq:max_div_ml}
\max_{\bphi} \frac{\mathcal{D}\left[ u_{\bphi}(\bx) \parallel  p^{(i)}(\bx)  \right ]}{\mathcal{D}\left[ u_{\bphi}(\bx) \parallel \sum_{i\neq j} \omega_{j} p^{(j)}(\bx) \right ]}
\end{equation}
provided $\sum_{i\neq j} \omega_j = 1$. This optimization can be either made jointly for every subdataset, or by sampling the task set $\mathcal{T}$. This formulation is a typical \textit{meta-learning objective} \citet{hospedales2020meta}
\begin{equation}
\label{eq:meta-obj}
\min_\Omega \mathbb{E}_{\mathcal{T} \sim p(\mathcal{T})} \mathcal{L}(\mathcal{D}, \Omega)
\end{equation}
where $\Omega$ is the across-task knowledge (or meta-knowledge), as for example the mixture coefficients $\omega_i \in \mathbb{R}$, and $\mathcal{L}(\cdot)$ is the meta-objective. Meta-learning setups allow us to split the training in two different optimization levels (called \textit{episodes}), that can be trained jointly or alternatively: the \textit{inner} task, generally a causual learning procedure (for instance fitting a single dataset), and the \textit{outer} task, handling the optimization of obj.~\ref{eq:meta-obj} over the set of datasets with meta-parameters $\Omega$.\\
Using the meta-learning objective (\ref{eq:meta-obj}) for training generative models may allow us to give the system additional insights about its own generation, that has been claimed as a mandatory property for creative artificial agents \citet{bundy1994difference, buchanan2001creativity}. Hence, detaching the outer loop from the inner loop would then enable to separate domain-specific optimization issues from global generation heuristics, that can be then designed separately (for example by dynamically adapting data groupings, as in curriculum learning \citet{soviany2022curriculum}). Furthermore, meta-learning approaches are known to conveniently deal with limited amounts of data, providing interesting solutions for zero-/few-shot learning \citet{jamal2019task, chen2020new, finn2017model}. This can be important in musical co-creation, where continual learning \citet{von2019continual, duan2017one} could be used for real-time learning in interactive setups. Finally, such approaches could also allow us to control the typicality of the generated data by reinforcing outliers or dynamically creating new datasets, drifting away from full data-centered approaches to the notion of \textit{inspiring set} \citet{rusu2018meta}.\\

\textbf{Diverging from high-level composition rules.} The weighted sum in equation~\ref{eq:max_div_ml} represents a mixture of multiple datasets, hence a composition of dataset in the data domain. Similarly to sec.~\ref{sec:gen_max_div}, we can extend obj.~\ref{eq:max_div_ml} to the parameter space to make our target model diverge from mixtures of parameters. We now take a collection of latent generative models  $p^{(i)}_{\btheta}(\bx | \bz)$, each one having a stacked architecture of $L$ layers with parameters $\{ \btheta^{(i)}_l\}_{l=1...L}, \ \btheta_l \in \Theta_l$, and trained on a different dataset $\bX^{(i)}$. We can also perform the maximum divergence meta-objective eq.~\ref{eq:max_div_ml} on the posterior parameter distributions : 
\begin{equation}
\label{eq:max_div_ml_layers}
\max_{\bphi} \sum_{l \in \Lambda}^L \frac{\mathcal{D}\left[  u(\bphi_l |\bx, \bz) \parallel p^{(i)}(\btheta_l |\bx, \bz) \right ]}{\mathcal{D}\left[  u(\bphi_l |\bx, \bz) \parallel \sum_{i\neq j} \omega_{j} p^{(j)}(\btheta_l |\bx, \bz) \right ]}
\end{equation}
where $\Lambda$ can be the full layer range $[1, ..., L]$, or a specific set of layers. The mixtures of weights $\sum_{i \neq j} \omega_{i} p^{(i)}(\btheta_l)$ are generalizations of \textit{model interpolations}, already investigated in model transfer and fine-tuning methods, that we may consider as machine learning equivalent of high-level composition rules \citet{epstein2020interpolating}. Hence, we can make the distribution $u(\bphi_l | \bx, \bz)$ diverge from parameter mixtures in specific layers of information, allowing us to learn or adjust the influence of each sub-dataset by including $\{ \omega_i \}_{i\in \mathbb{N}}$, and/or incorporating the set $\Lambda$ as a meta-parameter in the outer loop. 

\vspace{-0.2cm}

\section{Links to computational creativity}
\label{sec:cc}

The domain of computational creativity (CC) focuses on the adaptation of state-search problems to creative systems \citet{wiggins2006searching}. However, such approaches use task-centered computational algorithms to solve non-unique and context-dependent objectives, imposing to reformulate open tasks as closed objectives. While CC originally focuses on autonomous generative algorithms, most concepts can be interestingly translated to the use of machine-learning models in co-creative setups \citet{franceschelli2021creativity, esling2020creativity, berns2020bridging}. In this section, we briefly review how our proposed formulation can be understood within this field, and how it could be used towards a creative evaluation of these systems.

\subsection{Composition, exploration, transformation}
In her seminal work, Margaret Boden \citet{boden1998creativity} models CC as three different operations on a given conceptual space: \textit{combination, exploration}, and \textit{transformation}. Conceptual spaces are closely linked to \textit{generative factors} and \textit{disentanglement}, which are the overarching goals of generative models since their onset. Although ML methods could be considered as typical \textit{knowledge-based} creative systems, we can question here what would be their corresponding conceptual spaces. Considering that we cannot consider $\bx$ as conceptual, as it is our data domain, the main candidates are the latent space $\bz$ that can be considered the \textit{knowledge representation}, and the parameter space $\btheta$, that can be considered the \textit{state} of our model. We develop this analysis across different spaces in the following paragraphs.

\paragraph{Wiggins' creative systems.} An important framework for analyzing creative systems was proposed by Wiggins, defining a conceptual space $\mathcal{C}$ included in a super-set $\mathcal{U}$ \citet{wiggins2006preliminary}. A creative system is then defined by a set of restriction rules $\mathcal{R}$, restricting $\mathcal{U}$ to $\mathcal{C}$, and a set of transformation rules $\mathcal{M}$, transforming $\mathcal{C}$ to novel conceptual spaces, and an evaluation function $\mathcal{E}$, that can be used by the model to evaluate its own generations. In our case, $\mathcal{R}$ would be the inspiring sets $\bx$ or $p_{\btheta}(\bx)$, and the transformation rule is the optimization setup that we target with our proposed objectives \ref{eq:div_max} and \ref{eq:max_div_2}.

\textbf{Compositionality.} Compositionality in generative models is a widely studied task, which can be addressed in different manners. One approach aims to give explicit semantics to individual dimensions of the latent space $\mathbf{z}$, that are made orthogonal through \textit{disentanglement} metrics (as claimed by the xAI domain \citet{llano2020explainable, bryan2021exploring}). However, there is no proof of the ability of the model to generate novel content in unexplored semantic combinations. The other approach, coming from \textit{active divergence}, rather combines layers of different fine-tuned models to generate cross-domain images, as discussed in sec.~\ref{sec:meta-max}. Hence, in the first case, the conceptual space is the set of individual dimensions $z_i$, while in the second it is the set of parameters $\btheta^{(i)}$ for different datasets. 

\textbf{Explorative.} The most straightforward way of exploring generative models is, if available, to roam the latent space $\mathbf{z}$ whether by hand, by sampling, or by automatic routines as shown sec.~\ref{sec:div_max} \citet{fernandes2020evolutionary, cherti2017out}. Oppositely, exploration in the parameter space is generally difficult because of the number of weights in modern ML models. Hence, in this case, the conceptual space would ideally be $\bz$.

\textbf{Transformative.} While transformative creativity is a slightly more peculiar to define in generative models, explorative and transformative creativities can be linked by the existence of the super-set $\mathcal{U}$ \cite{wiggins2006preliminary}. Hence, identifying $\mathcal{U}$ as the space $\mathcal{Z} \times \Theta$, a machine learning system with transformative creativity would be able to morph a distribution $u(\bz, \btheta | \bx)$ relying on meta-learning optimization schemes, hence decoupling the concept space $\mathcal{C}$ in the inner loop and the transformative operator defined by the outer loop.

\textbf{H-creativity and P-creativity.} 
Another interesting separation made by Boden is between "P-creativity" (at which point the generated content is original for the model itself) and "H-creativity" (if the generated content is new to all existing instances of the domain). In our case, we can identify P-creativity as how the algorithm can diverge from its knowledge, hence directly corresponding to the divergence proposed in our objectives. On the contrary, H-creativity is much harder to quantize as it is essentially dependent to the cultural and social context of the generated context.

\subsection{CC-inspired evaluations of machine-learning generative models}

While creative evaluation of generative systems is a highly non-trivial tasks, several frameworks have been proposed to evaluate these models on selected auxiliary criteria \citet{colton2008creativity, eigenfeldt2012evaluating, jordanous2012standardised}. Besides allowing to analyze and eventually compare creativity-oriented models (\textit{summative evaluation}), it can also be used involved in the design process  (\textit{formative evaluation})  \citet{jordanous2012standardised}. In this subsection, we will discuss how we can bridge such criteria with the maximization framework described in sec.~\ref{sec:formulation}, eventually allowing to integrate such formative evaluation in the learning process. \vspace{-0.6cm}

\paragraph{On value, typicality and novelty.} 
According to some CC researchers an important property of creative systems is their ability to evaluate their own generations, in order to promote \textit{creative autonomy} \citet{bundy1994difference, buchanan2001creativity}. Conceptually, Ritchie \& al. \cite{ritchie2007some} proposed to evaluate these models on three different properties: \textit{novelty, value}, and \textit{surprise}, that can be used by the system to evaluate its own generations. However, the quantization of these computational measures is a challenging endeavour, notably for value. Traditional ML algorithms collapse the evaluation function $\mathcal{E}$ as being the same as the training loss. In most cases, this also encompasses the measure of \textit{typicality} (that is here the inverse of novelty). Hence, computationally addressing the notion of value is much more complicated, as this notion is inherently culturally- and socially-dependent. Yet, the lack of creativity of ML generative models could be explained by the fact that these algorithms are trained to identify typicality and value, preventing them to diverge from their knowledge. For this reason, some attempts to overcome this gap by using specialized discriminators as loss functions \citet{elgammal2015quantifying, franceschelli2022deepcreativity} provide interesting avenues for \textit{culturally}-informed models, and hence H-creativity. However, such methods do not really differentiate these two measures and require an important amount of annotated data. \\
Regarding our proposition, we can see that obj.~\ref{eq:div_max} and obj.~\ref{eq:max_div_2} generalize attempts made so far to promote diversity and extrapolation in generative systems, possibly at the expense of value. Furthermore, the meta-learning objective~\ref{eq:max_div_2} allows to differentiate optimization in the search space $\mathcal{Z}$ and the state space (parameter space $\Theta$), proposing a further step towards self-supervised generation and creativity at the meta-level \citet{buchanan2001creativity}. Regarding creative evaluations, we can see that obj.~\ref{eq:div_max} and \ref{eq:max_div_2} aim to maximize the divergence between the generations and the inspiring sets, directly providing an evaluation of the system's \textit{novelty}. Moreover, this optimization only assumes the computability of the chosen divergence $\mathcal{D}$, and is independent of the chosen architecture. We think that evaluating \textit{value} has to be done with complementary criteria that must be designed for specific applications, and may either consists in 1) developing domain-specific quality measures for the generated content, or 2) be context-dependant discriminators, as performed in \citet{elgammal2015quantifying}. 
\vspace{-0.5cm}
\paragraph{On surprise.} The notion of \textit{surprise}, considered by Boden as an important quality of creative systems, is also difficult to accurately delineate. Indeed, it can encompass a notion of statistical unexpectedness, a criterion over the user/audience reception, and an expected behavior of the creative system \citet{franceschelli2021creativity}. Regarding the first aspect, links to information theory have been made through the notion of \textit{Bayesian surprise}, allowing to locate zones of posterior distributions that drift away from the prior assumptions of the model \citet{baldi2010bits}. The second approach have been explored through composer-audience architectures by Bunescu \& al. \citet{bunescu2019learning} where a composer model is given the expectations of an audience model to produce its own future inputs. However, such models are only able to perform sequential generation. Our position is that evaluating surprise in ML-based generative models needs to differentiate these complementary aspects. The first aspect of surprise concerns how the inner stochasticity (latent distributions, noise modulation) of a given model acts on the generated content. It could measure the output variability obtained by sampling the stochastic component $\bepsilon$, or the ratio between the variability of changing either the latent variable $\bz$ or $\bepsilon$. It could also be obtained through measuring the entropy 
$\mathbb{H} \big[ u_{\bphi}(\bx; \epsilon) ] \big]$, giving a global measure on how $\bepsilon$ acts on the maximum divergence objective. Regarding the second aspect, composer-audience architectures can provide fruitful computational ways to measure and force the variability of the model, and stimulate their use in co-creative setups ; however, this may also slow down the appropriation process of their users.


\section{Discussion}
\label{sec:discussion}
\textbf{Machine learning and computational creativity.}
Building bridges between ML-based generative models and creativity is non-trivial for several reasons. First, as argued in \citet{ritchie2007some}, \textit{creativity} in the ordinary use is considered as "natural" and based on human behaviour. With this definition creativity is not only impossible to reach for artificial systems but also gets stuck into anthropomorphism, banishing the possible developments of an alternative artificial creativity to barren lands. Conversely, introducing the notion of creativity in machine learning is difficult, as explicitly designing losses for creativity is an uphill battle, such that such systems focus most of the time on narrow but precise and well-defined tasks (for instance most generative systems for audio still focus on the problem of text-to-speech approaches). Because of this difficulty, computational creativity aims to unveil abstract concepts about creative processes rather than developing specific measures, that we have to adapt to the machine learning domain if we want to turn generative models to creative systems. \\
Hence, we think that creative evaluations of these systems must not only be adapted but should also embrace and reflect the diversity of their possible applications. Our contribution attempted to provide theoretical solutions to deal with the problem of \textit{extensional cases}, by dissecting both opportunities and limitations offered by the redirecting traditional ML setups to creativity-oriented objectives. We saw that formulating extrapolation as a maximum divergence optimization is raising questions that are directly addressed by computational creativity: \textit{non-uniqueness} of solutions; value versus typicality; composition, transformation and exploration. We also argued that machine learning systems with transformative creativity could only be possible using meta-learning schemes, also allowing to considerably reduce the amount of needed data and then shift from \textit{training data} to \textit{inspiring sets}. We think that the proposed objectives could both provide promising ways to quantitatively measure how the generated distribution diverge from the actual data distribution, and how to incorporate such constraints directly in the training process. In the latter case, these constraints may then be optimized directly (assuming the development of adequate restrictions), or as an indirect regularization for latent or parametric distributions; moreover, this optimization could also be made either at initialization, or by cloning pre-trained models. Alternatively, these quantitative criteria could be used as measurements in more general creative frameworks, such as SPECS evaluation \cite{jordanous2012standardised}. While these objectives remain very general, we expect the resulting behaviour of the generated content to change drastically across models, and we leave to future works technical solutions of applying this new framework to specific architectures and creative purposes.

\textbf{Co-creation and social impacts.} Another important point of analysing ML-based generative models through computational creativity also allow us to evaluate the \textit{specific} benefits of bringing AI to human-computer co-creative setups. Indeed, Wiggins rightly objects that the \textit{value} of an artifact is also dependent from what we expect from its generator \citet{wiggins2006preliminary}, obfuscating furthermore what we could expect from such models. Indeed, as described by the notion of \textit{latent heat effect} in computational creativity, the creative responsibility given to a system does not necessarily increase the value given to its input \citet{colton2008creativity, colton2012computational}. This fact is even more true with machine learning methods, as the alliance between computational power and data-based approaches can amplify the evaluation biases between creators, users, and audience. Hence, properly evaluating creative generative models requires an active collaboration between scientific, design and artistic communities, as claimed by example in musical meta-creation setups \citet{eigenfeldt2012evaluating}. This position is directly addressed by the recent introduction of \textit{active divergence}, proposing motley ways to make pre-trained generative models diverge from their original distribution \citet{broad2021active}. However, most of these methods are still globally restricted to post-training / fine-tuning heuristics. While the contribution of these practice-driven approaches \citet{mccormack2020design} are substantial for our objectives, most of them do not provide quantitative ways to integrate directly these objectives in the training process. Directly addressing \textit{extrapolation} during training with \textit{guided} divergence would be a further step towards modern issues of AI creative systems: systems opened to edition and hijacking, embeddable into human/computer co-creation frameworks, fully employing the high capacity of modern machine learning systems, and fostering the system to generate added value rather than reproduction (for both artistic interest and copyrighting issues). Furthermore, we think that the development of creativity-oriented machine learning in artistic communities can provide a precious experimentation ground for more general purposes \citet{bryan2021exploring}, and make a further step towards a fruitful and ethical collaboration between humans and AI. While we know that our contributions are so far theoretical and deprived of tangible examples (that we will investigate in future works), we hope that our propositions may inspire and motivate efforts and discussion towards this direction. 

\textbf{Specific directions for musical creation.} As the presented methods are quite general, we will shortly review some specific aspects for musical creation. While some methods presented below specifically address symbolical music generation, such as composer-audience architectures, few attempts have been made to directly address the extrapolation problem in the musical or audio domain. Indeed, besides a network bending attempt on the DDSP model \citet{engel2020ddsp, yee2021studio} few active divergence methods have been yet investigated, neither in symbolical nor audio domain\footnote{see first author musical contribution at AIMC2022, \textit{aletheia}.}. However, such approaches could provide significant improvements in musical co-creative setups. In the symbolical domain, as discussed sec.~\ref{sec:meta-max}, the application of maximum divergence to few-shot and/or continual learning could greatly help such models to be applied in real-time improvisation setups and musical agents \cite{tatar2019musical, nika2017dyci2}, allowing them to quickly adapt to the musical context while going away from usual repetitive or imitative behaviours. In the signal domain, such methods could allow to generate content based on existing signals, that are typically very hard to learn, but to diverge in specific manners that could motivate artists to explore neural audio synthesis more extensively. However, the musical domain has specific aspects that have still to be precisely addressed: notably, the dynamical aspect of generation. Indeed, the variety of strategies to address temporality in ML audio generative models may prevent to design general frameworks to address extrapolation: some models intend to generate local chunks of data, some others to generate single samples, and some others to generate full songs. Hence, developing adequate extrapolation strategies will not only depend of the generation scope, but also to the desired interaction with the interested musicians, performers, and sound designers. \vspace{-0.2cm}

\section{Conclusion}
\vspace{-0.2cm} In this article, we provided prospective and theoretical propositions to bend usual ML optimization setups towards novel objectives to encourage generative models to diverge from an existing dataset, rather than only focusing on reproduction. We derived some important properties and limitations of the proposed objective, extended it to other aspects of latent generative models, and proposed a meta-learning framework to control specific aspects of the divergence. Then we proposed several bridges and directions between the proposed objectives and notions coming from computational creativity, hoping that such links could help the integration of creative ML generative models in co-creative setups. In further works, we intend to develop specific audio generative models implementing the proposed objectives, and to propose creative evaluations that could exhibit the relationships between the models and existing concepts of computational creativity. 

\bibliography{main}
\bibliographystyle{abbrvnat}

\end{document}